\documentclass[runningheads,a4paper]{llncs}

\usepackage{amssymb}
\usepackage{graphicx}

\usepackage{balance}  
\usepackage{graphics} 
\usepackage{times}    
\usepackage{url}      
\usepackage{graphicx}
\usepackage{array,multirow}
\usepackage{longtable}
\usepackage{todonotes}
\usepackage{tikz}
\usetikzlibrary{positioning}

\usepackage{url}
\urldef{\mailsa}\path|{rdgain, jialin.liu,sml,dperez}@essex.ac.uk|  

\newcommand{\keywords}[1]{\par\addvspace\baselineskip
\noindent\keywordname\enspace\ignorespaces#1}

\begin{document}

\mainmatter  

\title{Analysis of Vanilla Rolling Horizon Evolution Parameters in General Video Game Playing}

\titlerunning{Analysis of Vanilla RHEA Parameters in GVGP}

%
%
\author{Raluca D. Gaina, Jialin Liu, Simon M. Lucas, Diego P\'erez-Li\'ebana}
%

\institute{School of Computer Science and Electronic Engineering, \\University of Essex, Colchester CO4 3SQ, UK\\
\mailsa\\
}

%
%

\toctitle{Lecture Notes in Computer Science}
\tocauthor{Authors' Instructions}
\maketitle

\begin{abstract}
Monte Carlo Tree Search techniques have generally dominated General Video Game Playing, but recent research has started looking at Evolutionary Algorithms and their potential at matching Tree Search level of play or even outperforming these methods. Online or Rolling Horizon Evolution is one of the options available to evolve sequences of actions for planning in General Video Game Playing, but no research has been done up to date that explores the capabilities of the vanilla version of this algorithm in multiple games. This study aims to critically analyse the different configurations regarding population size and individual length in a set of $20$ games from the General Video Game AI corpus. Distinctions are made between deterministic and stochastic games, and the implications of using superior time budgets are studied. Results show that there is scope for the use of these techniques, which in some configurations outperform Monte Carlo Tree Search, and also suggest that further research in these methods could boost their performance.
\end{abstract}


\section{Introduction}

General Video Game Playing (GVGP) is a sub-domain of Artificial General Intelligence (AGI), which aims to create an agent capable of achieving a high level of play in any given environment, that was potentially previously unknown. It uses video games as testbeds for this purpose because of their complex nature, offering practical problems in a constrained environment where it is easy to quantify results and observe performance. In contrast with other domains such as robotics, where errors are expensive to correct, video games are cheap alternatives for testing AI algorithms, as well as having the possibility of multiple tests run very quickly (due to modern computational power).

The General Video Game AI Competition (GVGAI) \cite{perez2016general,GVGAI} offers a large corpus of games described in a plain text language, making it easy to run general AI agents in several different environments and analyse their performance. The competition has already completed three editions of its single player track (starting in 2014), with two additional tracks running in 2016 for two player games~\cite{GVGAI2P} and level generation~\cite{GVGAILG}. Therefore, it is attracting a large interest on an international scale, with close to a hundred participants every year across its different tracks. 

This competition is becoming a popular way of benchmarking AI algorithms such as enforced hill climbing \cite{hillclimbing}, algorithms employing advanced path finding or using the knowledge gained during the game in interesting ways \cite{Perez2014,Chu2015}, or dominant Monte Carlo Tree Search techniques \cite{Park2015}. All of the authors appear to agree on the complexity of the problem proposed, as well as its importance, going beyond the realm of video games towards that of AGI.

Among the techniques employed over the last years of the GVGAI, one of the most promising is that of Rolling Horizon Evolutionary Algorithms (RHEA). These methods, rather than basing the search on game tree structures, use influences from biological sciences to evolve a population of individuals until a suitable one, corresponding to a solution to the problem, is obtained. The way they are applied to the domain of GVGP is by encoding sequences of in-game actions as individuals, using  heuristics to analyse the value of each sequence~\cite{Perez2015}.

Up to date, there is no in depth evaluation of the vanilla version of RHEA on the GVGAI framework, attending to certain crucial parameters such as population size and individual length. It is hardly possible that the same parameter setting would work equally well for all of the assorted games of the GVGAI corpus: on one hand, these games can vary in many forms, such as their level of stochasticity, average duration of a game, presence or absence of other NPCs, etc, but on the other hand, variations of the population size and the lengths of the action sequences explored may be sensitive to variations in the game design space.

The first objective of this paper is to perform an analysis of the vanilla version of RHEA (see Section~\ref{ssec:ea}) on a subset of $20$ GVGAI games, with special focus on the population size and the individual length of this technique. This analysis is performed attending to the different games presented, and their stochastic nature. Additionally, this study aims to make a comparison with the sample Open Loop Monte Carlo Tree Search (OLMCTS), the best sample agent included in the GVGAI framework, which is actually the starting point of several winners of the competition in past editions.

The rest of this paper is structured as follows: Section~\ref{sec:research} reviews work already present in the literature on this topic, with Section~\ref{sec:background} detailing background information on the framework and algorithms used. Section~\ref{sec:experiments} describes the approach taken and the experimental setup, while Section~\ref{sec:results} presents the results obtained from this experiment. The paper concludes in Section~\ref{sec:conclusion} with a discussion of the results and notes on future work that will be undertaken as a consequence of this study.

\section{Relevant Research}\label{sec:research}

The popularity of General Game Playing (GGP) has increased in the last decade, since M. Genesereth et al.~\cite{Genesereth2005} organised the first GGP competition allowing participants to submit game agents to play in a diverse collection of board games. Sharma et al.~\cite{Sharma2009} motivates research in this area by bringing to attention how agents trained without prior knowledge of the game and excelling in specific games, such as TD-Gammon in Backgammon \cite{Tesauro1995} and Blondie24 in Checkers \cite{Blondie24}, cannot be successfully applied in other scenarios or environments. 

The problem is further expanded to video games in General Video Game Playing (GVGP~\cite{GVGP}), which provide the agents with new and possibly more complex challenges due to a higher and continuous, in practice, rate of actions. One of the first frameworks to allow testing of such general agents was the Arcade Learning Environment (ALE)~\cite{bellemare13arcade}, later used as benchmark for applying Deep Q-Learning to achieve human level of play on the Atari 2600 collection~\cite{mnih-dqn-2015}. The way the world was presented to the agents in this framework was via screen capture; they would return an action to be performed and the next game state would be processed by the system. 

 
Monte Carlo Tree Search methods have dominated GVGP so far, and their variations have been explored in various works~\cite{MCTSsurvey}. However, Evolutionary Algorithms (EA) show great promise at obtaining just as good, if not better, performance. Perez et al. \cite{Perez2013} compare EA techniques with tree search on the Physical Salesman Travelling Problem, and their results are satisfactory, encouraging research in the area. In their work, the authors employ several techniques to improve the state evaluation function, such as avoiding opposite actions, movement blocks and pheromone exploration.

Samothrakis et al. \cite{Samothrakis2013} compare two variations of the Rolling Horizon setting of EAs in a number of continuous environments, including a Lunar Lander game. The first algorithm uses a co-variance matrix, while the second employs a value optimisation algorithm. The Rolling Horizon refers to evolving plans of actions and, at each game step, executing the first action that appears to be the best at present, while starting fresh and creating a new plan for the next move, sequentially increasing the "horizon". Their research suggests EAs to be viable algorithms in general environments, and that a deeper exploration should be performed with an emphasis on heuristic improvement.

N. Justesen et al.~\cite{justesen2016online} used online evolution for action decision in Hero Academy, a game in which each player counts on multiple units to move in a single turn, presenting a branching factor of a million actions. In this study, groups of actions are evolved for a single turn, to be performed by up to $6$ different units. With a fixed population of $100$ individuals, the authors show that online evolution is able to beat MCTS and other greedy methods. Later, Wang et al.~\cite{wang2016portfolio} employed a modified version of online evolution using a portfolio of script to play Starcraft micro. In this work, rather than evolving groups or sequences of actions, the algorithm evolved plans to determine which script (among a set of available ones) each unit should use at each time step. Each gene in the individual represents a script that will be executed by a given unit in the next turn.

Other different approaches to EAs have been explored in the past, such as combining them with other techniques in order to produce hybrids, and take advantage of the benefits of each algorithm \cite{Horn2016}. For example, evolution was used during the simulation phase in a Monte Carlo Tree Search algorithm by Perez et al.~\cite{Perez2014}, or, for a different effect, the MCTS parameters were adjusted with evolutionary methods~\cite{Lucas2014}. There has been recent work that has attempted to give more focus to the evolutionary process and instead integrates tree structures into EAs, or uses N-armed bandit techniques and Upper Confidence Bounds (UCB) for informing and guiding the evolution process~\cite{BanditRMHC}.

\section{Background}\label{sec:background}

\subsection{The GVGAI Framework}

The experiments presented in this paper were run within the General Video Game AI framework\footnote{www.gvgai.net}, frequently used in recent literature for benchmarking Artificial Intelligence agents due to its large and constantly increasing collection of games. This framework currently includes $100$ single player and $40$ two-player games, of both deterministic and stochastic nature. All of them are real time games, where the agents receive a $1$ second time budget for initialisation purposes and a $40ms$ budget for selecting an action to be performed during each game step.

The action space available to the agents is limited to a maximum of $5$, although it can vary across games. The agents may choose to perform no action (ACTION\_NIL; it is important to note that this is not equivalent to the avatar stopping movement), to move in a certain direction (ACTION\_LEFT or RIGHT, UP or DOWN, correspondingly), or to perform a special action (ACTION\_USE) that depends on the game, and may range from shooting to creating or activating various game objects. 

Concrete information about the game rules is not available to the agents, although they do have access to details about the current game state through a State Observation object. This includes the current score, game tick, a description of the state of the avatar (such as position, orientation, resources etc.), and data about other game objects (such as NPCs, portals or static objects).

Another tool available to the agents through this framework is a Forward Model (FM), which allows for simulation of possible future states of the game (this simulated state may not be accurate in stochastic games). In order to advance the Forward Model, the agent must supply one of the legal actions of the game to an \textit{advance} function, which would roll the state of the game forward following this move.

\begin{figure}[!t]\label{fig:games}
\centering
\includegraphics[width=.49\textwidth]{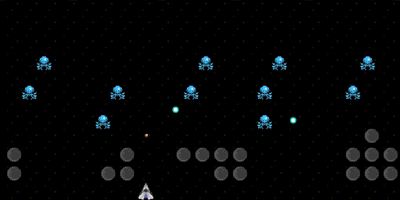}
\includegraphics[width=.49\textwidth]{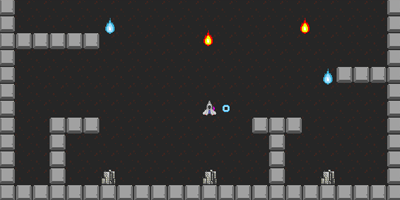}\\
\includegraphics[width=.49\textwidth]{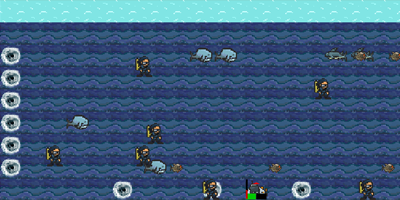}
\includegraphics[width=.49\textwidth]{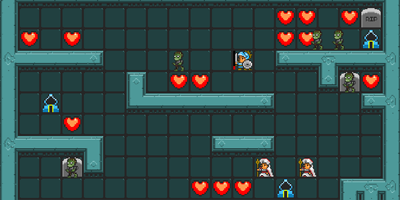}
\caption{Games in GVGAI Framework: Aliens, Missile Command, Sea Quest and Survive Zombies (from left corner, clockwise).}
\label{fig:gvgaiGames}
\end{figure}

Games vary in nature not only in their probabilistic states, but also in the presence of certain game objects (e.g. NPCs and portals), scoring methods (binary, in which $1$ point is awarded for winning, $0$ otherwise; incremental, which sees continuous small rewards spread out in the game; or discontinuous, in which certain actions or sequences of actions may produce a sudden large gain), or the conditions which lead to an end state (e.g. counters, timers or exit doors). This results in a great variety of games, which truly tests the abilities of general agents. Figure~\ref{fig:gvgaiGames} shows a few examples of games included in this framework, which were also employed in this study.

The ranking of controllers in the GVGAI competition used for the results analysis of this paper employs a Formula 1 point system per game: agents are sorted based on their performance (win percentage, score and time steps, in this order, with the secondary ones used as tiebreakers if needed) for each game, then awarded a number of points depending on their position: $25$ for the first, then $18$, $15$, $12$, $10$, $8$, $6$, $4$, $2$, $1$ and $0$ for all subsequent entries. The points are then summed to a total used to determine the position in the overall rankings. This system is meant to emphasise the generic aspect of the competition, as achieving a high average win rate is not equivalent to performing well across all games. 

\subsection{Rolling Horizon Evolutionary Algorithms} \label{ssec:ea}

Rolling Horizon Evolutionary Algorithms (RHEA)~\cite{Perez2013} are a subset of EAs which use populations of individuals representing action plans or sequences of actions. The individuals are evaluated by simulating moves ahead using a Forward Model. From the current state of the games, all actions (genes of the individual) are executed in order, until a terminal state or the length of the individual is reached. The state reached at that point is then evaluated with a heuristic function and the value assigned as the fitness of the individual.

\begin{figure}[!t]
\centering
\includegraphics[width=0.75\columnwidth]{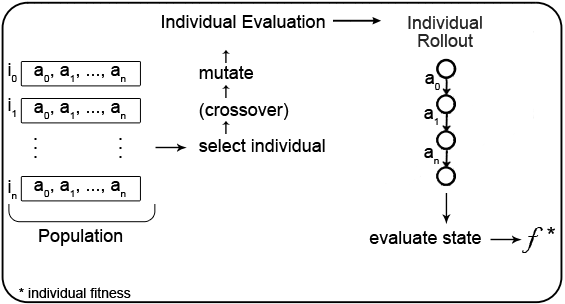}
\caption{Rolling Horizon Evolutionary Algorithm cycle.}
\label{fig:rhsteps}
\end{figure}

In general, the algorithm starts with a random population of individuals. At each game step it applies traditional genetic operators (such as mutation, randomly changing some actions in the sequence, and cross-over, combining individuals in different ways) to obtain new individuals for the next generation of the population. Each one of them is then evaluated and assigned a fitness, according to which the population is sorted and only the best are carried forward to subsequent generations. This process ends when an end condition is satisfied, such as a time or memory limit reached or a certain number of iterations have been performed. The action selected by the algorithm is represented by the first gene in the best individual found at the end of the evolutionary process. The action is played in the game, a new state is received in the next step by the agent, and new iterations are performed to evolve new action plans.

As the agents have a limited amount of time to make decisions in real-time games, one of the popular methods in the literature consists of generating only one new individual at each generation, therefore making it possible to interrupt the process at any point. The most basic form this algorithm can take is that of a Random Mutation Hill Climber \cite{Mitchell1998}, where the population size is only $1$, using the mutation operator as the only way to navigate through the search space. 

\subsection{Open Loop Monte Carlo Tree Search (OLMCTS)} \label{ssec:olmcts}

\begin{figure}[!t]
\centering
\caption[ ]{\label{fig:mcts} Monte Carlo Tree Search steps~\cite{MCTSsurvey}}
\includegraphics[width=.8\textwidth]{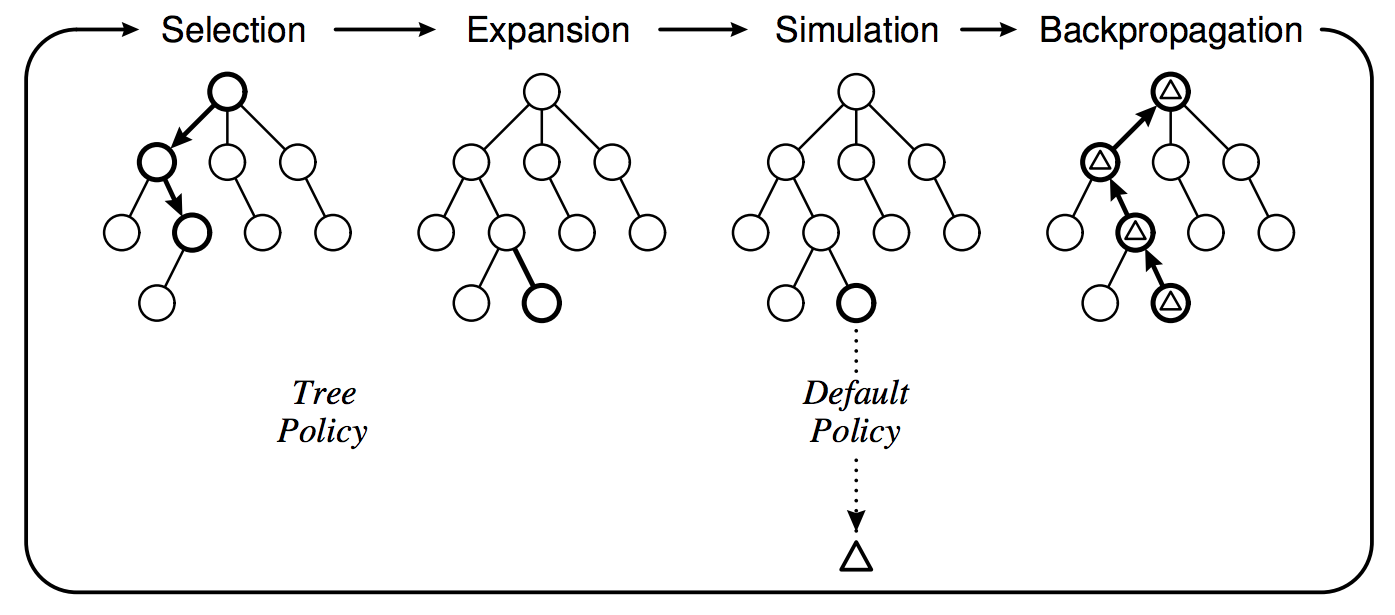}
\end{figure}

Open Loop Monte Carlo Tree Search (OLMCTS) is an MCTS implementation for the GVGAI framework. This particular agent does not store the states of the game in the nodes of the tree, but instead uses the forward model to reevaluate each action. OLMCTS uses four simple steps to produce a high level of play: selection (using a tree policy to select one of the current leaves of the tree, which is not yet fully expanded), expansion (adding a new child of the selected node to the tree), simulation (a Monte Carlo process using the forward model to advance through the game with random actions) and back-propagation (the state reached after the MC simulation is evaluated using a heuristic and its value backed up the tree to the root node, updating all other parent nodes). The steps of the MCTS algorithm are depicted in Figure~\ref{fig:mcts}.

When reaching the limit of its execution budget (memory, time, iterations, or, as is the case of this paper, number of calls to the forward model \textit{advance} function), the algorithm returns action to apply via a recommendation policy. In the GVGAI implementation of this agent, the action returned is that of the child of the root node that has been selected more often. For an in depth description of Monte Carlo Tree Search, variants, improvements, and applications, the reader is referred to~\cite{MCTSsurvey}.

\section{Approach and Experimental Setup}\label{sec:experiments}

\subsection{Methods}

This paper analyses how modifying the population size ($P$) and individual length ($L$) configuration of the vanilla Rolling Horizon Evolutionary Algorithm (RHEA) impacts performance in a generic setting. Exhaustive experiments were run on all combinations between population sizes $P=\{1, 2, 5, 7, 10, 13, 20\}$ and individual lengths $L=\{6, 8, 10, 12, 14, 16, 20\}$. The budget defined for planning at each game step was set as $480$ Forward Model calls to the \textit{advance} function, the average number of calls OLMCTS is able to perform in $40ms$ of thinking time in the games of this framework\footnote{Using these forward model calls instead of real execution time is more robust to fluctuations on the machine used to run the experiments, making it time independent and results comparable across different architectures.}. Larger values for either individual length or population size were not considered due to the limited budget and the complete nature of the experiment (analysis of all combinations); values above $24$ would not allow in certain cases for a full evaluation of even one population.

The fitness function used by RHEA evaluates the state reached after executing the sequence of actions in an individual, and returns the current in-game score of the player. In the case where an end-game state has been reached, it instead gives a large penalty for losing the game (or, alternatively, a high reward for winning). 

To expand the analysis of the results, a particular configuration was also tested, using $P=24$ and $L=20$. Effectively, given the budget of $480$ Forward Model calls, this is an equivalent method of Random Search (RS). The algorithm only has enough budget to initialise the population before applying any genetic operator. In essence, this configuration evaluates $24$ random walks and returns the first action of the best sequence of moves found. 

The algorithm itself begins with the initialisation of the population, which sets each individual to a sequence of actions selected uniformly at random. The genes of the individual take integer values in the interval [0, N-1], where N is the number of available actions in that particular game state, therefore each value corresponding to an in-game legal action. The evolutionary process then proceeds in a slightly different way depending on the population size. For the case in which there is only one individual in a population, one new individual is mutated at each iteration and it replaces the first if its fitness is higher (RHEA is set to maximize the fitness provided by the value function). 

For a population of size $2$, the best individual is passed on to the next generation unchanged (elitism of $1$), then uniform crossover and mutation are applied to the $2$ individuals to generate the second solution for the new population. If the population contains $3$ or more individuals, similar rules apply, but the $2$ parents are selected for crossover through a tournament of size $2$. The mutation operator always modifies one gene of the individual, chosen uniformly at random. It is important to note that the initialisation is counted in the budget received for evolution, in order to ensure that there is a trade-off in higher population sizes.

In order to validate the results, Open Loop Monte Carlo Tree Search was also tested on the same set of $20$ games, under the same budget conditions. OLMCTS has proven to be the dominating technique out of the sample ones provided in the GVGAI competition, with numerous participants using it as a basis for their entries before adding various enhancements on top of its vanilla form. The winner of the first edition of the competition in 2014, Adrien Cou\"etoux~\cite{GVGAI}, employed an Open Loop technique quite similar to this algorithm.

\subsection{Games}

All of the combinations explored in this study were run on $20$ games of the GVGAI corpus, on all $5$ levels, $20$ times each, resulting in $100$ games played per configuration. The games were selected using two different classifications present in literature in order to balance the game set and analyse performance on an assorted selection of different games. The first classification was that generated by Mark Nelson~\cite{MCTSScaling:CIG16} in his analysis of the vanilla Monte Carlo Tree Search algorithm in $62$ of the games in the framework, sorted using the win rate of MCTS as a simple criterion. The second classification considered for this study was the clustering of $49$ games by Bontrager et al.~\cite{Bontrager2016}, which separated the games into groups based on their similarity in terms of game features. Combining these two lists and uniformly sampling from both provided a diverse subset appropriate for this experiment, which contains $10$ stochastic and $10$ deterministic games. See Table~\ref{tab:games} for the name of these games and the indices used in later figures in this document.

\begin{table}[!t]
\begin{center}
\caption{Names, indexes and types of the $20$ games from the subset selected. Legend: S - Stochastic, D - Deterministic.}
\begin{tabular}{|c|>{\centering\arraybackslash} m{1.5cm}|c||c|>{\centering\arraybackslash} m{1.5cm}|c||c|>{\centering\arraybackslash} m{1.5cm}|c||c|>{\centering\arraybackslash} m{1.5cm}|c|}
\hline
\textbf{Idx} & \textbf{Name}  & \textbf{Type} & \textbf{Idx} & \textbf{Name}  & \textbf{Type} & \textbf{Idx} & \textbf{Name} & \textbf{Type} & \textbf{Idx} & \textbf{Name}  & \textbf{Type} \\
\hline\hline
$0$ & Aliens & S & $4$ & Bait & D & $13$ & Butterflies & S & $15$ & Camel Race & D \\
\hline
$18$ & Chase & D & $22$ & Chopper & S & $25$ & Crossfire & S & $29$ & Dig Dug & S \\
\hline
$36$ & Escape & D & $46$ & Hungry Birds & D & $49$ & Infection & S & $50$ & Intersection & S \\
\hline
$58$ & Lemmings & D & $60$ & Missile Command & D & $61$ & Modality & D & $67$ & Plaque Attack & D \\
\hline
$75$ & Roguelike & S & $77$ & Sea Quest & S & $84$ & Survive Zombies & S & $91$ & Wait for Breakfast & D \\
\hline
\end{tabular}
\label{tab:games}
\end{center}
\end{table}

\section{Results and Discussion}\label{sec:results}

This section presents and analyses the results obtained from different angles. Observations are made attending to the nature of the game and variations of the population size and individual length. Section~\ref{ssec:pop} compares the performance using smaller or larger population, while Section~\ref{ssec:len} discusses the impact of individual length. Later, the performance of RHEA is also compared to RS employing different budgets (Section~\ref{ssec:rs}) and OLMCTS (Section~\ref{ssec:olmcts}) as supplied by the GVGAI framework. As the game set used is divided equally between deterministic and stochastic games, an in-depth analysis is carried out on each game type, although it is not implied the trend would carry through in other games of the same type.

Additionally, a Mann-Whitney non-parametric test was used to measure the statistical significance of results for each game ($p$-value $=$ $0.05$). Table~\ref{tab:avgData} summarises the winning rates of all configurations tested in this study.

\begin{table*}[!h]
\begin{center}
\caption{Winning rate for different values of population size ($P$) and individual length ($L$), in all $20$ tested games. Average of standard errors indicated between brackets. Highlighted in bold style is the best result.}
\begin{tabular}{|c|c|c|c|c|c|c|c|} 
\hline
 \textbf{P} & \textbf{L=6}  & \textbf{L=8}  & \textbf{L=10}  & \textbf{L=12}  & \textbf{L=14}  & \textbf{L=16}  & \textbf{L=20}  \\
\hline
 \textbf{1} & $35.45 (2.54)$  & $38.25 (2.54)$ & $37.95 (2.47)$ & $36.70 (2.58)$ & $34.20 (2.42)$ & $33.55 (2.57)$ & $33.15 (2.60) $ \\
\hline
 \textbf{2}   & $39.95 (2.62)$ & $40.95 (2.55)$ & $41.05 (2.62)$ & $40.25 (2.48)$ & $39.50 (2.56)$ & $38.75 (2.56)$ & $36.80 (2.60) $ \\
\hline
 \textbf{5}   & $42.55 (2.57)$ & $43.50 (2.39)$ & $44.65 (2.40)$ & $44.25 (2.38)$ & $43.80 (2.34)$ & $44.95 (2.53)$ & $46.05 (2.54) $ \\
\hline
 \textbf{7}   & $43.00 (2.49)$ & $42.60 (2.43)$ & $44.65 (2.36)$ & $44.35 (2.45)$ & $45.30 (2.23)$ & $44.80 (2.47)$ & $47.05 (2.56) $ \\
\hline
 \textbf{10}   & $42.25 (2.53)$ & $43.60 (2.49)$ & $44.05 (2.26)$ & $45.80 (2.47)$ & $45.05 (2.35)$ & $46.60 (2.45)$ & $46.80 (2.49) $ \\
\hline
 \textbf{13}   & $42.65 (2.43)$ & $45.15 (2.48)$ & $45.15 (2.47)$ & $45.00 (2.42)$ & $46.25 (2.41)$ & $47.40 (2.30)$ & $47.05 (2.42)$ \\
\hline
 \textbf{20}   & $42.75 (2.51)$ & $43.20 (2.60)$ & $44.75 (2.31)$ & $45.50 (2.34)$ & $46.45 (2.32)$ & $46.30 (2.32)$ & $\textbf{47.50 (2.33)}$  \\
\hline
\end{tabular}
\label{tab:avgData}
\end{center}

\begin{center}
\caption{Winning rate for different values of population size ($P$) and individual length ($L$), in the $10$ deterministic tested games. Average of standard errors indicated between brackets. Highlighted in bold style is the best result.}
\begin{tabular}{|c|c|c|c|c|c|c|c|} 
\hline
 \textbf{P} & \textbf{L=6}  & \textbf{L=8}  & \textbf{L=10}  & \textbf{L=12}  & \textbf{L=14}  & \textbf{L=16}  & \textbf{L=20}  \\
\hline
 \textbf{1}   & $22.30 (2.88)$ & $26.80 (2.95)$ & $26.90 (2.93)$ & $25.30 (2.91)$ & $24.20 (2.84)$ & $23.00 (3.01)$ & $22.50 (2.99) $ \\
\hline
 \textbf{2}   & $26.40 (3.13)$ & $26.80 (3.08)$ & $27.90 (3.05)$ & $27.90 (2.92)$ & $27.10 (2.91)$ & $26.80 (2.93)$ & $24.50 (2.99) $ \\
\hline
 \textbf{5}   & $28.70 (3.08)$ & $29.70 (3.10)$ & $31.90 (3.18)$ & $31.80 (2.88)$ & $30.00 (2.86)$ & $32.00 (3.04)$ & $32.20 (3.19) $ \\
\hline
 \textbf{7}   & $28.80 (3.26)$ & $29.00 (3.00)$ & $30.80 (3.09)$ & $30.40 (3.01)$ & $31.70 (2.82)$ & $32.00 (2.99)$ & $34.30 (3.12) $ \\
\hline
 \textbf{10}   & $27.70 (3.18)$ & $31.00 (3.27)$ & $29.50 (2.90)$ & $33.00 (3.03)$ & $32.60 (2.94)$ & $32.40 (3.11)$ & $33.20 (3.05) $ \\
\hline
 \textbf{13}   & $28.90 (3.19)$ & $32.20 (3.32)$ & $32.10 (3.06)$ & $31.80 (3.07)$ & $33.30 (3.18)$ & $\textbf{34.70 (2.88)}$ & $34.00 (2.97) $ \\
\hline
 \textbf{20}   & $28.60 (3.19)$ & $29.90 (3.34)$ & $31.50 (2.87)$ & $32.30 (3.05)$ & $33.10 (3.11)$ & $32.10 (2.84)$ & $34.30 (3.02) $ \\
\hline
\end{tabular}
\label{tab:avgData}
\end{center}

\begin{center}
\caption{Winning rate for different values of population size ($P$) and individual length ($L$), in the $10$ stochastic tested games. Average of standard errors indicated between brackets. Highlighted in bold style is the best result.}
\begin{tabular}{|c|c|c|c|c|c|c|c|} 
\hline
 \textbf{P} & \textbf{L=6}  & \textbf{L=8}  & \textbf{L=10}  & \textbf{L=12}  & \textbf{L=14}  & \textbf{L=16}  & \textbf{L=20}  \\
\hline
 \textbf{1}   & $48.60 (2.20)$ & $49.70 (2.13)$ & $49.00 (2.01)$ & $48.10 (2.25)$ & $44.20 (2.00)$ & $44.10 (2.12)$ & $43.80 (2.22) $ \\
\hline
 \textbf{2}   & $53.50 (2.12)$ & $55.10 (2.02)$ & $54.20 (2.20)$ & $52.60 (2.05)$ & $51.90 (2.20)$ & $50.70 (2.20)$ & $49.10 (2.22) $ \\
\hline
 \textbf{5}   & $56.40 (2.07)$ & $57.30 (1.68)$ & $57.40 (1.61)$ & $56.70 (1.88)$ & $57.60 (1.81)$ & $57.90 (2.01)$ & $59.90 (1.89) $ \\
\hline
 \textbf{7}   & $57.20 (1.72)$ & $56.20 (1.85)$ & $58.50 (1.64)$ & $58.30 (1.90)$ & $58.90 (1.63)$ & $57.60 (1.95)$ & $59.80 (2.00) $ \\
\hline
 \textbf{10}   & $56.80 (1.88)$ & $56.20 (1.71)$ & $58.60 (1.63)$ & $58.60 (1.91)$ & $57.50 (1.77)$ & $\textbf{60.80 (1.79)}$ & $60.40 (1.93) $ \\
\hline
 \textbf{13}   & $56.40 (1.68)$ & $58.10 (1.65)$ & $58.20 (1.88)$ & $58.20 (1.76)$ & $59.20 (1.63)$ & $60.10 (1.71)$ & $60.10 (1.86) $ \\
\hline
 \textbf{20}   & $56.90 (1.83)$ & $56.50 (1.86)$ & $58.00 (1.74)$ & $58.70 (1.64)$ & $59.80 (1.53)$ & $60.50 (1.80)$  & $60.70 (1.64) $ \\
\hline
\end{tabular}
\label{tab:avgData}
\end{center}
\end{table*}

\subsection{Population Variation}\label{ssec:pop}

\begin{figure}[!t]
\centering
\caption[ ]{\label{fig:length} Change in winning rate as population size increases, for individual lengths $L=6$ and $L=14$, in all games tested for this paper. The Standard Error is shown by the shaded boundary. Please refer to Table~\ref{tab:games} for the names of the game indexes presented here.}
\includegraphics[width=.49\textwidth]{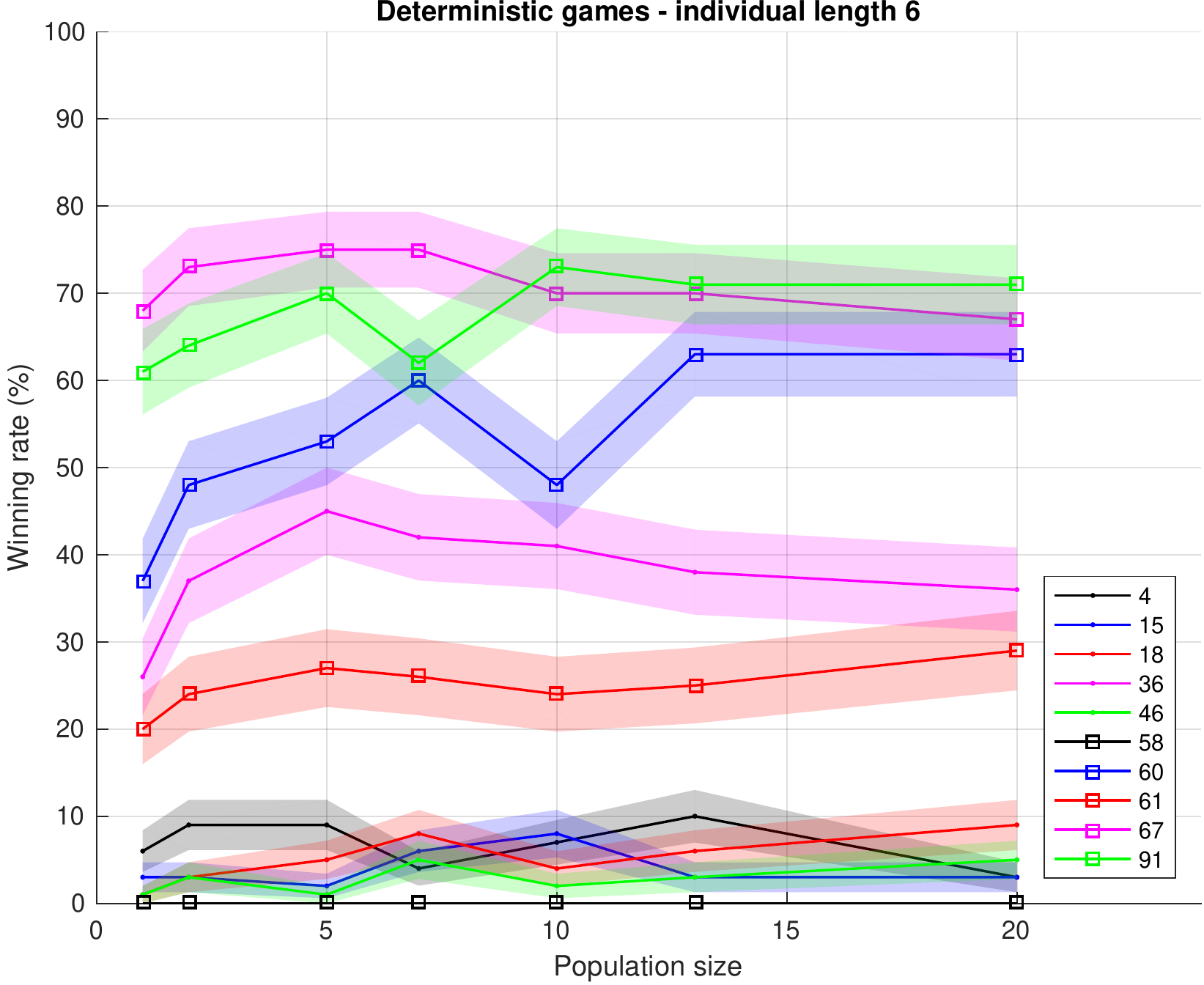}
\includegraphics[width=.49\textwidth]{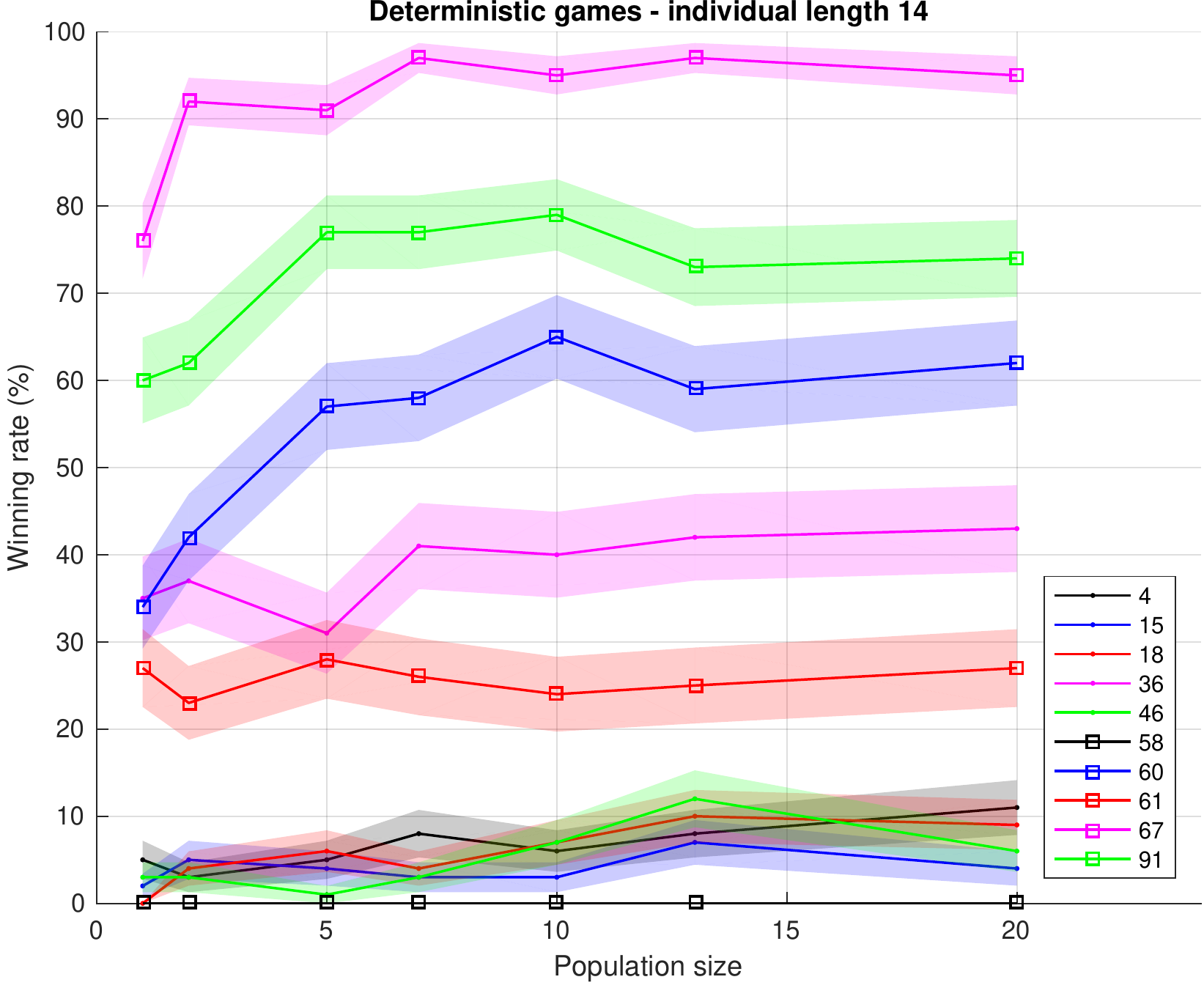}\\
\includegraphics[width=.49\textwidth]{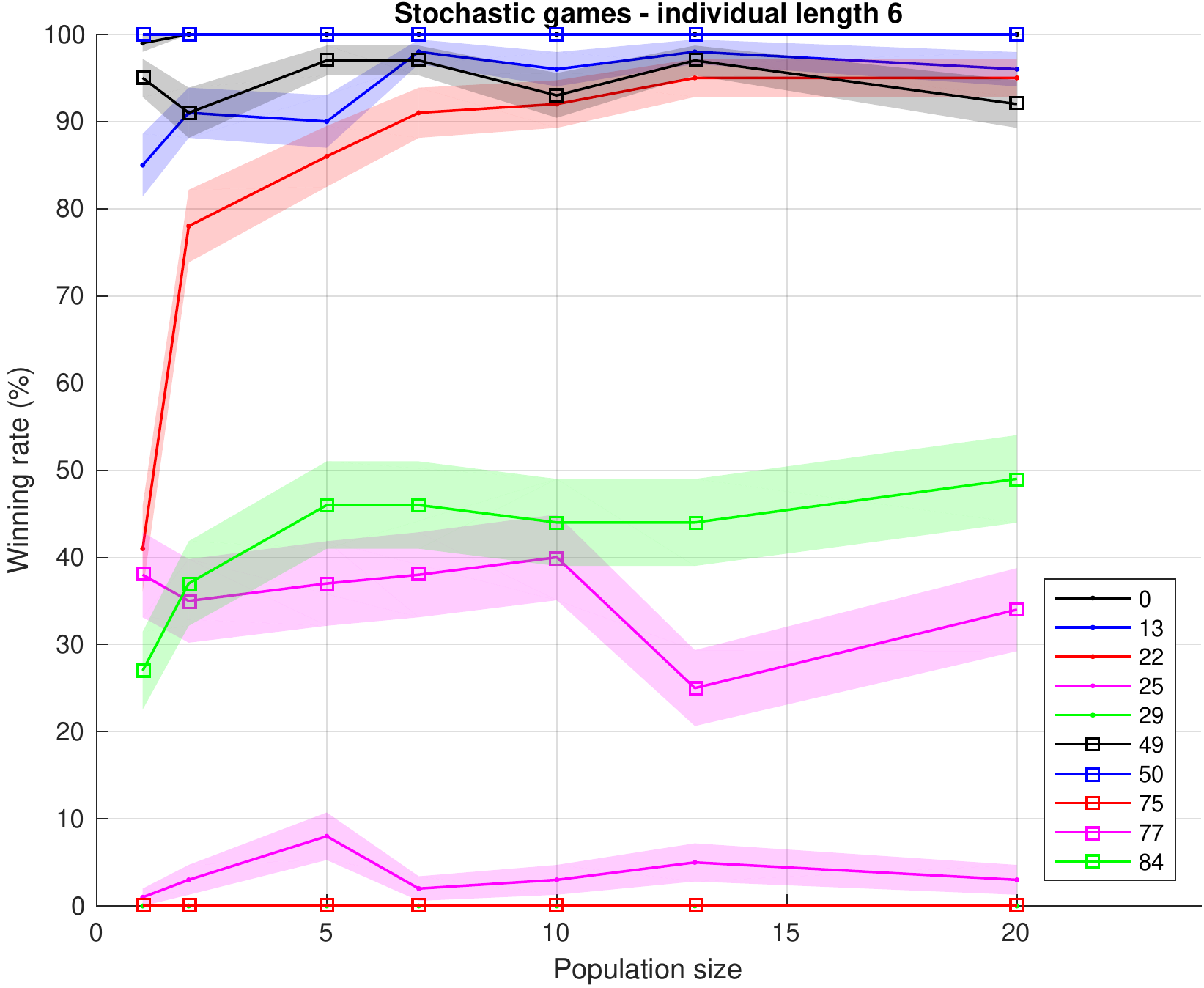}
\includegraphics[width=.49\textwidth]{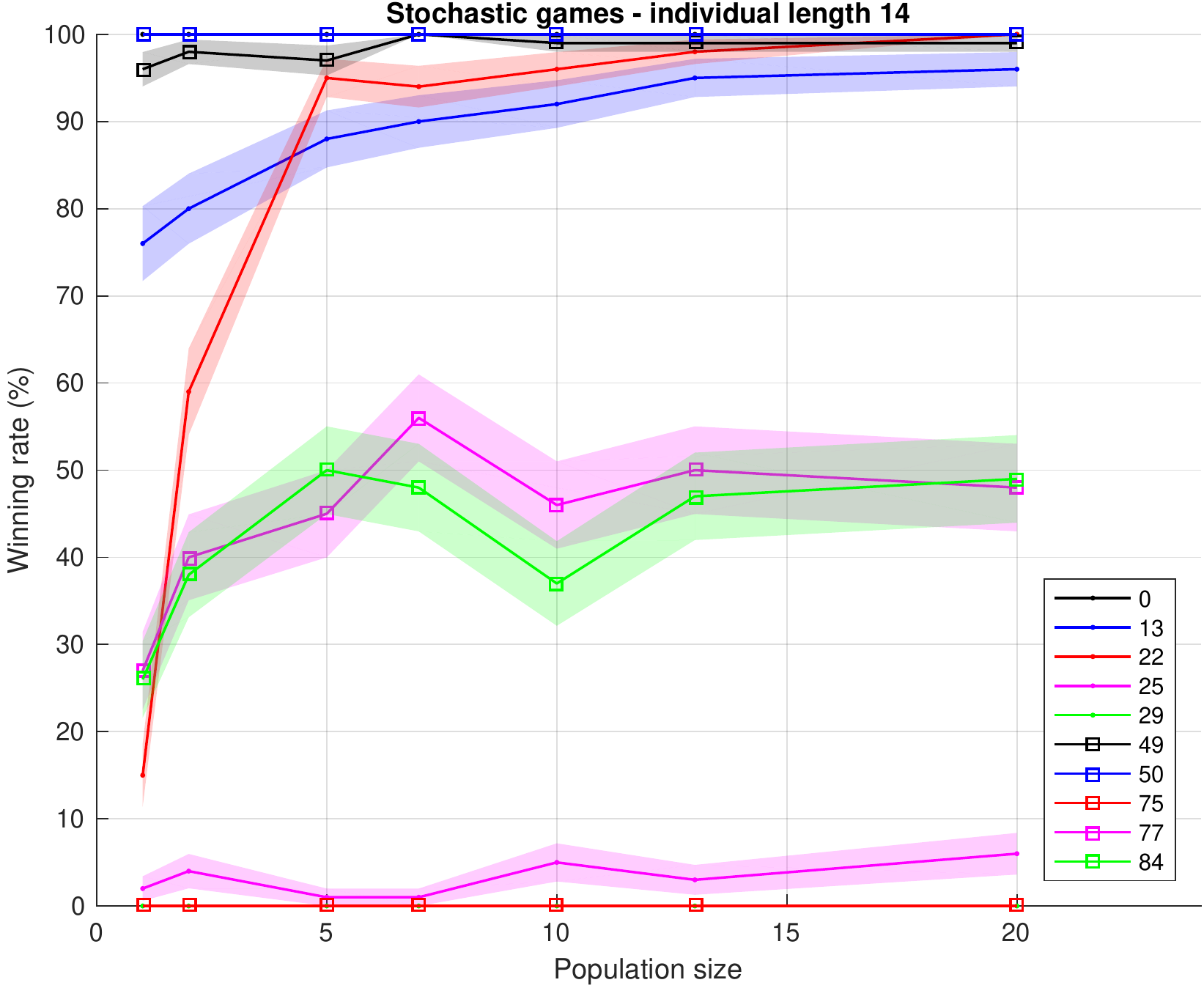}
\label{fig:pop}
\end{figure}

Figure~\ref{fig:pop} shows the change in winning rate as population size increases, for $L=6$ and $L=14$ (figures for other individual lengths have been omitted for the sake of space). 
Each of the $20$ games that these algorithm configurations were tested on showed different performance and variations. There is a trend noticed in most of the games, with win rate increasing, regardless of the game type (c.f. Table \ref{tab:avgData}). Exceptions are for games where the win rate starts at $100\%$, therefore leaving no room for improvement (games with indexes $0$ and $50$, \textit{Aliens} and \textit{Intersection}, respectively) or, on the contrary, when the win rate stays very close to $0\%$ due to outstanding difficulty (game index $75$, \textit{Roguelike}). The winning rate on game with index $25$, \textit{Crossfire}, increases significantly from $0$ to $10\%$ ($p$-value $=$ $0.02$) along with the increase in population size. This suggests that games which a priori seem unsolvable, can be approached by exploring more with a larger population.

\paragraph{Deterministic games} 
Winning rate increases progressively in most of the tested deterministic games  (Figure~\ref{fig:pop}, top). A high diversity of the performance over the tested games is observed, with the concrete winning rate having a high dependency on the given game. The games with indexes $60$ and $91$ (\textit{Missile Command} and \textit{Wait for Breakfast}, respectively), stand out in these cases as they achieve a larger increase in performance, particularly with longer individuals. 

\paragraph{Stochastic games} 
Regarding stochastic games (Figure~\ref{fig:pop}, bottom) in particular, it is important to separate them based on their probabilistic elements and their impact on the outcome of the game. For example, the game with index $84$, \textit{Survive Zombies}, has numerous random NPCs and probabilistic spawn points for all object types, in contrast with game numbered $0$, \textit{Aliens}, where its stochastic nature comes only from the NPCs dropping bombs in irregular intervals. 

In games numbered $13$ and $22$ (\textit{Butterflies} and \textit{Chopper} respectively), a big improvement in terms of winning rate is observed by increasing the population size from $1$ (the case in which there is no tournament) to $5$, and this remains stable with larger populations.

When the length of the individual is fixed to a small value, increasing the population size is not beneficial in all cases, sometimes having the opposite effect and causing a drop in win rate (games with indexes $77$ and $84$, \textit{Sea Quest} and \textit{Survive Zombies}, respectively). On the contrary, the game with index $22$, \textit{Chopper}, sees a great improvement (from an average of $29\%$ in population size $P=1$ to $98\%$ in population size $P=20$, $p$-value $\ll$ $0.001$, for both win rate and scores achieved). 

In general, a conclusion that could be drawn from these experiments is that increasing the population size rarely hinders the agent to find good solutions. In fact, in some cases it makes the difference between a very poor and a very successful performance (from $29\%$ to $98\%$ in \textit{Chopper}). An explanation for this phenomena could be that the higher diversity in the population allows the algorithm to perform a better exploration of the search space.

\subsection{Individual Variation}\label{ssec:len}
Figure~\ref{fig:length} illustrates the change of the winning rate in each of the 20 games as individual length increases, for population sizes $P=1$ and $P=5$. The full results using a variety of population size and individual length are given in Table \ref{tab:avgData}. Using identical numbers of individuals when the population size is large ($P\geq5$) and increasing the individual length, i.e., simulation depth, leads to a growth of winning rate (c.f. Table \ref{tab:avgData}).

\begin{figure}[t]
\centering
\caption[ ]{\label{fig:length} Change of the winning rate as individual length increases, for population sizes $P=1$ and $P=5$, in all games tested for this paper. The standard error is shown by the shaded boundary. Please refer to Table~\ref{tab:games} for the names of the game indexes presented here.}
\includegraphics[width=.49\textwidth]{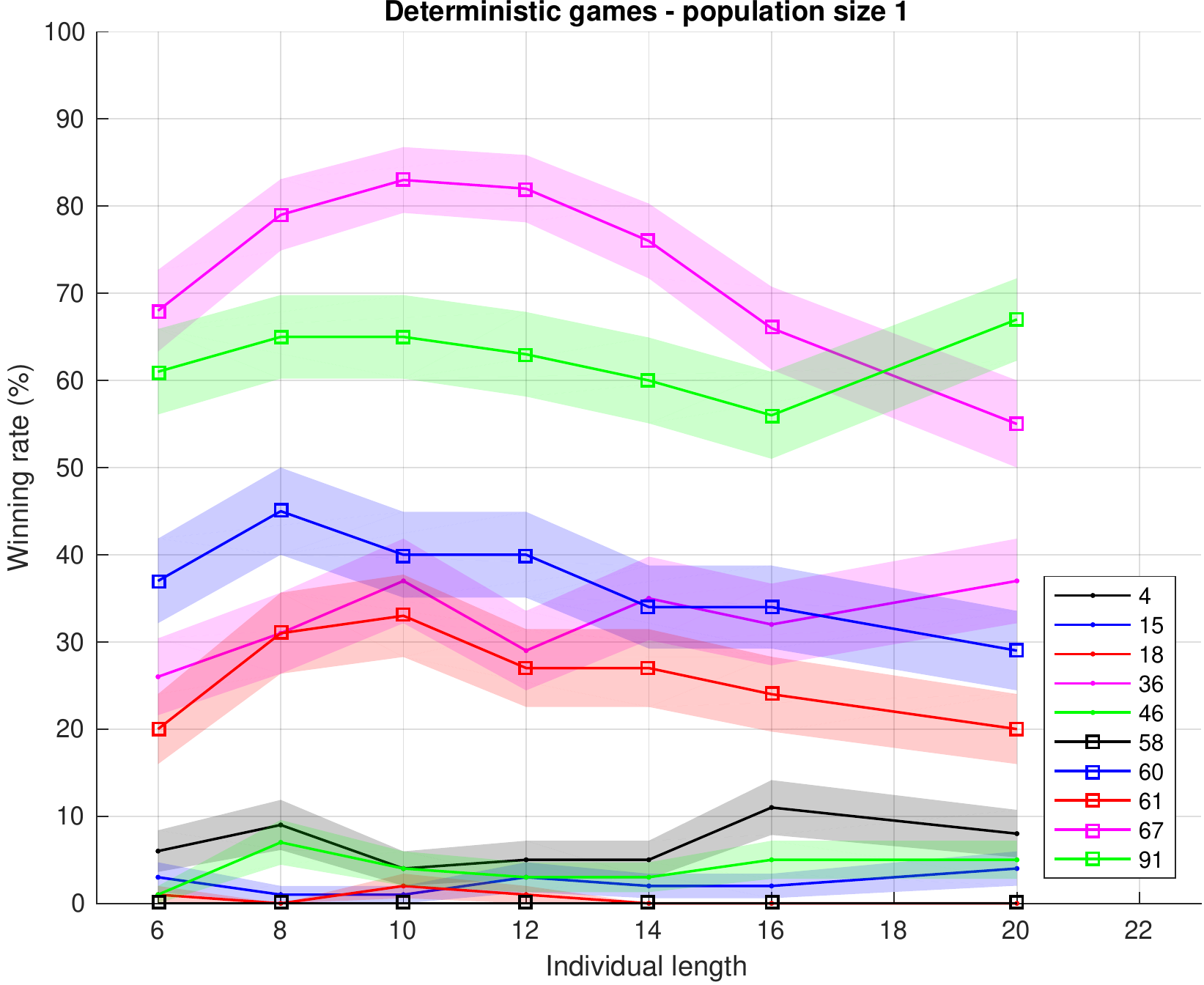}
\includegraphics[width=.49\textwidth]{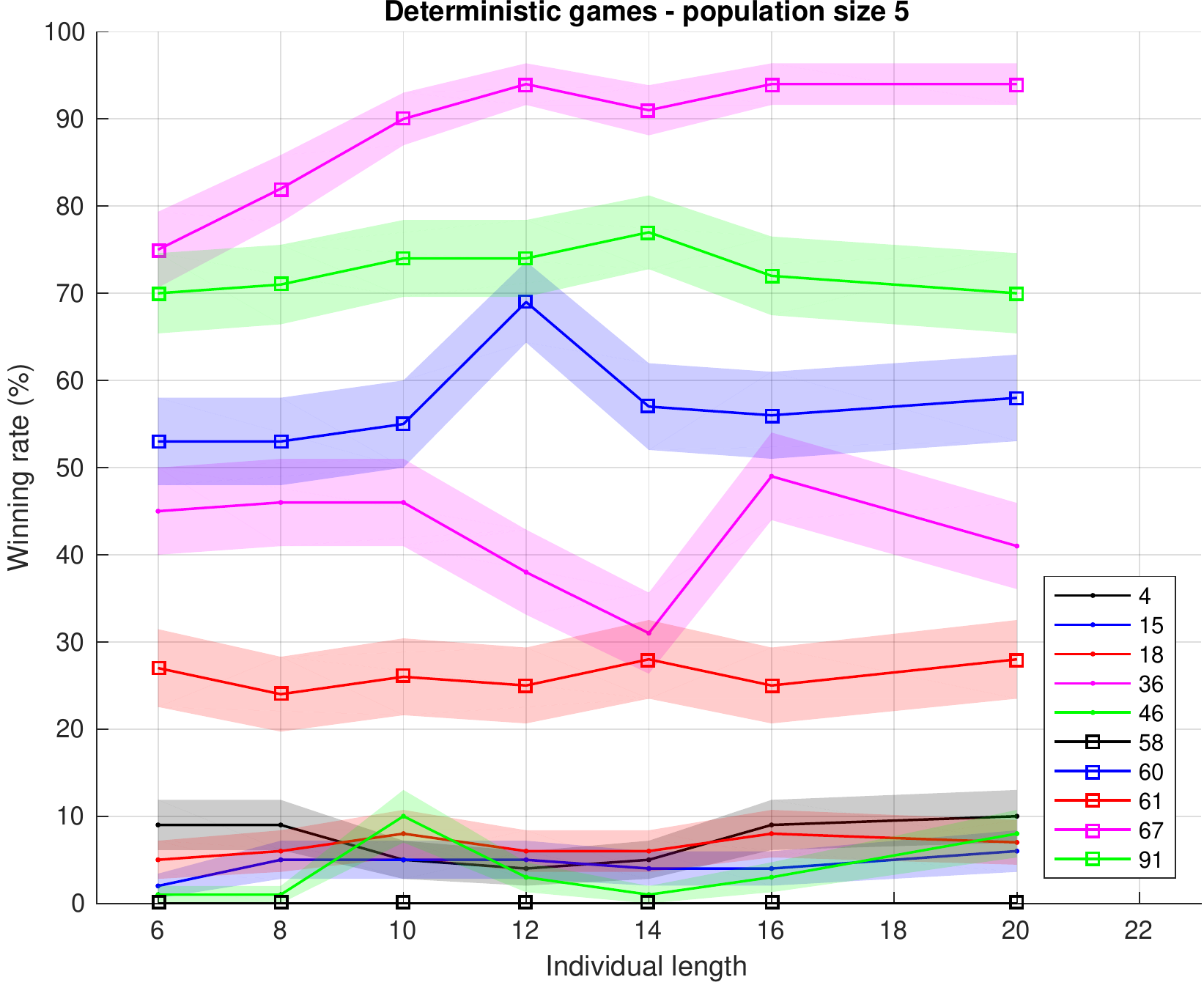}\\
\includegraphics[width=.49\textwidth]{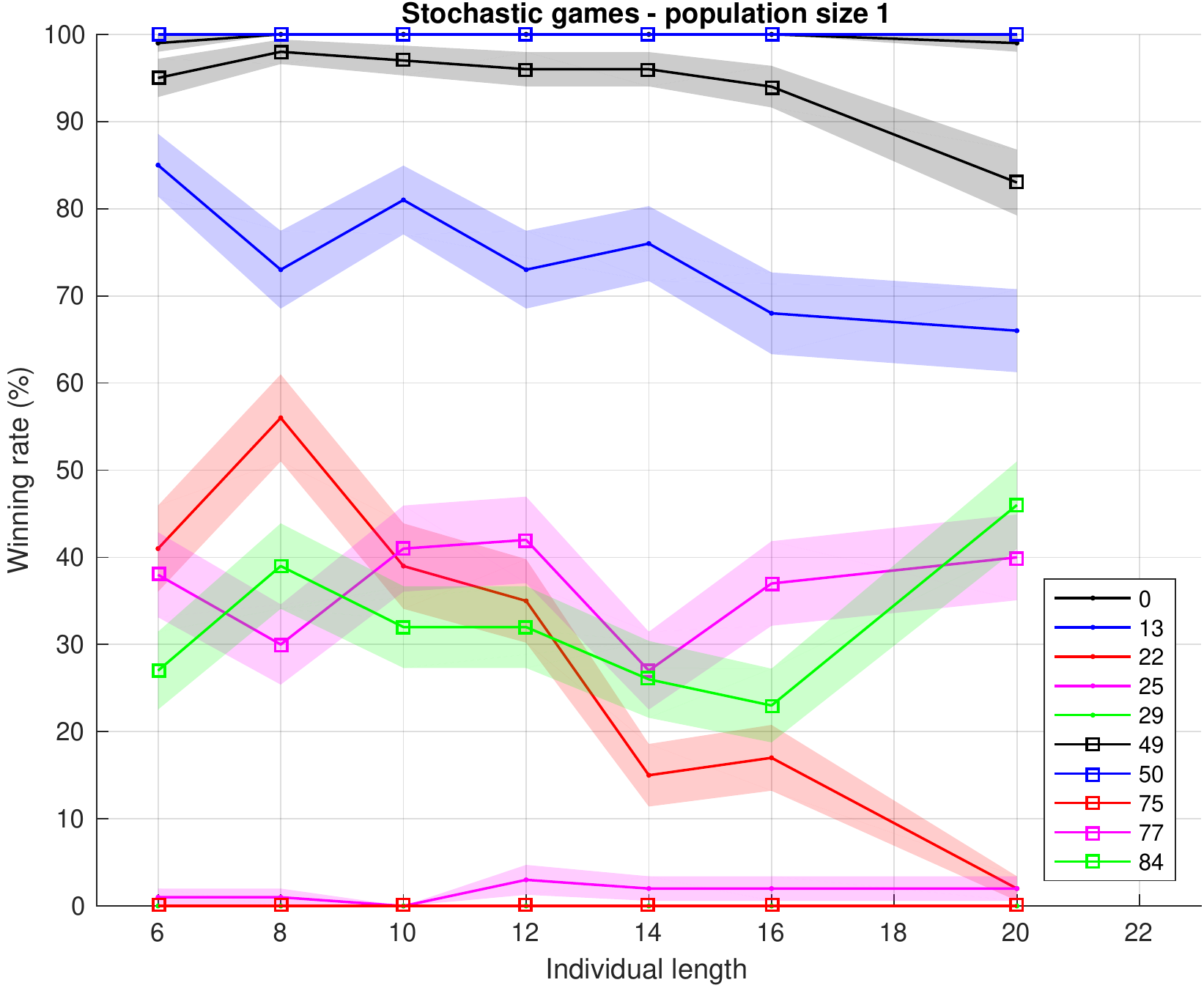}
\includegraphics[width=.49\textwidth]{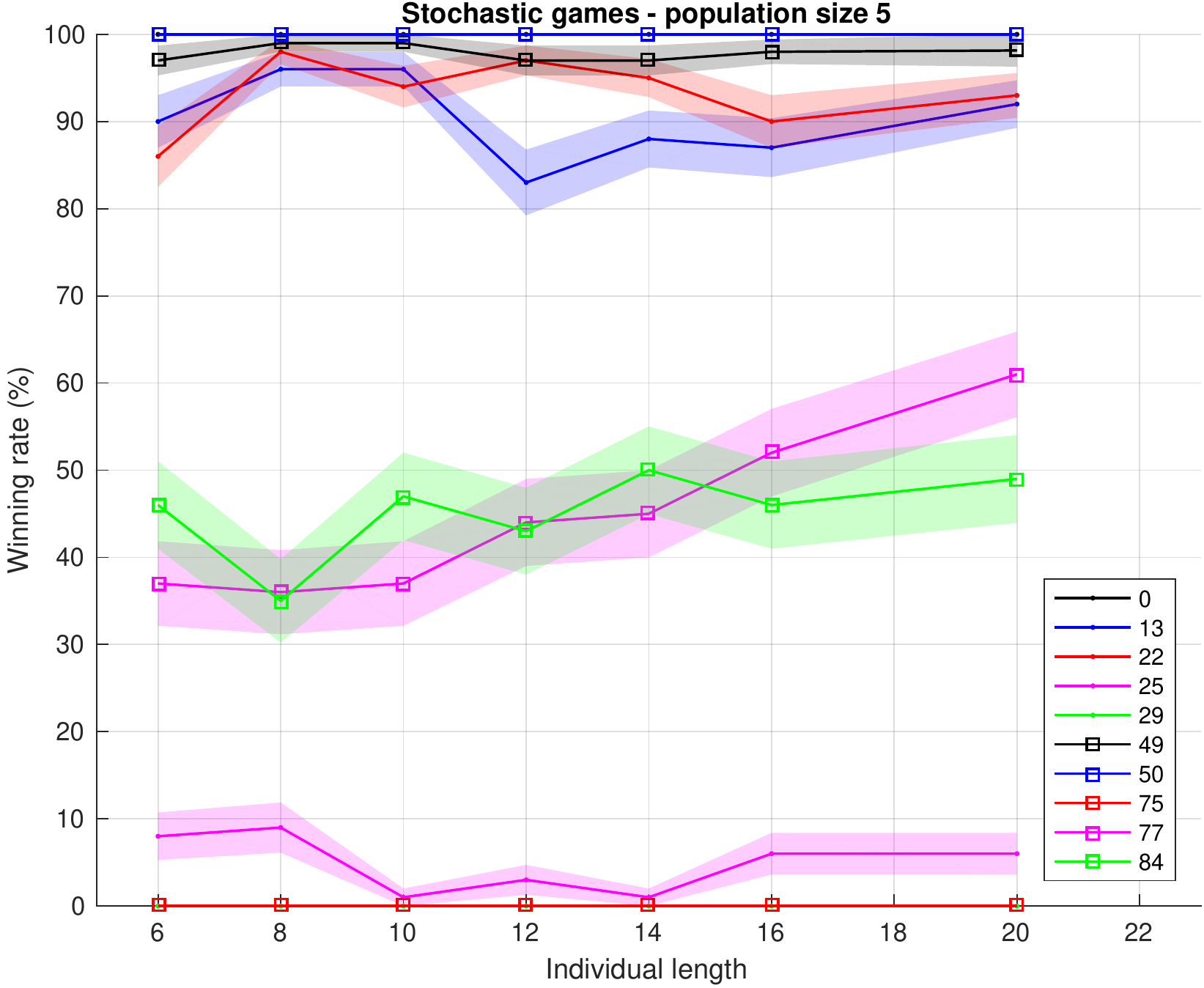}
\label{fig:ind}
\end{figure}

\paragraph{Deterministic games} 
When there is only one individual in the population, thus no crossover is involved, the winning rate experiences a significant increase followed by a drop along with the increase of individual length. This is due to the fact that the size of search space of solutions increases exponentially with the individual length. With few individuals evaluated, the algorithm struggles to find optimal solutions. 
This issue can be solved by increasing the population size, as shown in Figure~\ref{fig:length} (top).
For instance, the game with index $67$, \textit{Plaque Attack}, sees a variation from $68\%$ to $83$\% to $55\%$ with population size $P=1$; while with population size $P=5$, there is a constant increase from $79\%$ to $97\%$. 

\paragraph{Stochastic games} 
In stochastic games, however, matters are different.  In this case, the performance of the different variants of RHEA depends greatly on the game played. For instance, in game $13$ (\textit{Butterflies}), performance drops significantly ($p$-value = $0.001$) from a win rate of $91\%$ ($L=6$) to $75\%$ ($L=20$), using a population of $P=2$ individuals. An even bigger difference can be seen in game $22$ (\textit{Chopper}) which drops from $78\%$ ($L=6$) to $30\%$ ($L=20$) for a population of $P=2$ individuals ($p$-value $\ll$ $0.001$ for both win rate and in-game scores). No significant change in win rate can be appreciated in larger population sizes.

In general, increasing the length of the individual provides better solutions if the size of the population is high, although the effect of increasing the population size seems to be bigger. This can be clearly observed in the results reported in table~\ref{tab:avgData}.

\subsection{Random Search}\label{ssec:rs}
The version of RHEA using large values for population size and individual length is reminiscent of the Random Search (RS) algorithm. We perform a RS on the same set of games using $P=24$ individuals and simulation depth $L=20$. As a budget of $480$ calls to the forward model is allocated to this algorithm, RS is equivalent to RHEA using this population size and individual length. The average winning rate in each of the tested games is summarized in the last row of Table \ref{tab:rs}.

RS performs no worse than any variant of RHEA studied previously.
This result supports one of the main findings on this paper: the vanilla version of RHEA is not able to explore the search space better than (and, in most cases, not even as good as) RS in the framework tested when the budget is very limited. In order to test the limits and potential benefits of evolution, an additional set of experiments was run, using the same $P=24$, $L=20$ configuration, but increasing the forward model budget from $480$ \textit{advance} calls to $960$, $1440$ and $1920$. 
It's notable that, for these new budgets, the population is evolved during $2$, $3$ and $4$ generations, respectively.

\begin{table}[!t]
\begin{center}
\caption{Comparison of winning rates and points achieved by RHEA with different budgets and OLMCTS. It shows rates and points for all games (T), deterministic (D) and stochastic (S). 
With budget 480, the RS is equivalent to a RHEA using 24 individuals and individual length 20.}
\begin{tabular}{|c|>{\centering\arraybackslash} m{1.7cm}|>{\centering\arraybackslash} m{1.5cm}|>{\centering\arraybackslash} m{1.7cm}|>{\centering\arraybackslash} m{1.5cm}|>{\centering\arraybackslash} m{1.7cm}|>{\centering\arraybackslash} m{1.5cm}|}
\hline
\textbf{Algorithm} & \textbf{Average Wins (T)}  & \textbf{Points (T)} & \textbf{Average Wins (D)}  & \textbf{Points (D)} & \textbf{Average Wins (S)}  & \textbf{Points (S)} \\
\hline
 \textbf{RHEA-1920} & $48.25 (2.36) $ & $351$ & $36.30 (2.88)$ & $181$ & $60.20 (1.84)$ & $170$ \\
\hline
\textbf{RHEA-1440} & $ 48.05 (2.23) $ & $ 339$ & $35.40 (2.82)$ & $177$ & $60.70 (1.65)$ & $162$ \\
\hline
\textbf{RHEA-960} & $ 47.85 (2.39) $ & $323$ & $34.60 (2.99)$ & $162$ & $61.10 (1.79)$ & $161$ \\
\hline
\textbf{OLMCTS-480} & $ 41.45 (1.89) $ & $316$ & $22.20 (2.45)$ & $149$ & $60.70 (1.34)$ & $167$ \\
\hline
\textbf{RHEA/RS-480} & $46.60 (2.40) $ & $271$ & $32.90 (3.04)$ & $131$ & $60.30 (1.76)$ & $140$ \\
\hline
\end{tabular}
\label{tab:rs}
\end{center}
\end{table}

The results, presented in Table~\ref{tab:rs}, suggest that the solution recommended by RHEA at the end of optimisation converges towards the optimal solution while increasing the budget. As the budget becomes higher, the win rate increases first, to then stabilise when it reaches the highest budget tested. The difference observed is smaller than that given by the search in terms of population sizes and individual lengths.

In stochastic games, there is no difference observed in the average winning rate, but there is a small increase in ranking points, which vary according to the budget. However, there is a clearer improvement in performance distinguished in deterministic games. 
This may be due to the fact that resampling an individual is useless in deterministic games, whilst a single evaluation of a solution in a stochastic environment may be inaccurate.
  
\subsection{RHEA vs OLMCTS} \label{ssec:olmcts}

Table~\ref{tab:rs} also includes the performance of the GVGAI sample OLMCTS agent. 
The sample OLMCTS agent uses a playout depth of $10$, 
hence the comparisons presented here relate to RHEA configurations with individual length $L=10$. Results show that, although RHEA is significantly worse when its population size is small, it outperforms OLMCTS when the number of individuals per population is increased ($P > 5$). A second interesting contribution of this paper is that it is possible to create an RHEA capable of achieving a higher level of play than OLMCTS, which is the base of most dominating algorithms in the GVGAI literature.

In addition, OLMCTS also falls short when comparing it to RS with regards to the average percentage of victories. However, it does manage to gain a higher number of ranking points in these games against the other $4$ agents. Considering the fact that points are awarded for each game in order to value their generic capabilities, this result suggests that OLMCTS is more general than the vanilla version of RHEA.

Finally, if an analysis is carried out per game type, OLMCTS appears to be similar to RS in stochastic games but, not surprisingly, its performance is much worse than RS in deterministic games, becoming comparable to the worst configuration of RHEA found during these experiments (population size $P=1$ and individual length $L=20$).

\section{Conclusions and Future Work}\label{sec:conclusion}

This paper presents an analysis of population size and individual length of the vanilla version of Rolling Horizon Evolutionary Algorithm (RHEA). The performance of this algorithm is measured in terms of winning rate in a subset of $20$ games of the General Video Game AI corpus. These games were selected based on their difficulty and game features, in order to present a reduced set of challenges as assorted as possible. Games were also chosen so there would be a split between deterministic and stochastic ones.

One of the main findings of this research is the fact that RHEA is unable to find better solutions than Random Search (RS) in the settings explored, being worse than RS in many cases. 
Rather than an indication of RHEA being not suitable for GVGAI, these results suggest that the vanilla version of the algorithm is not able to explore the search space quickly enough given the limited budget. Therefore, this finding motivates research in RHEA, in order to find operators and techniques able to evolve sequences of actions in a more efficient way. The results presented in this paper with higher execution budgets are an indication that this is possible.

At the same time, this paper highlights another interesting conclusion: given the same length for the sequence of actions and the same budget ($480$ calls to the forward model), RHEA is able to outperform Open Loop Monte Carlo Tree Search (OLMCTS) when configured with a high population size. Most of the entries of the GVGAI competition, including some of the winners, base their entries in OLMCTS or similar tree search methods. Thus, RHEA presents itself as a valuable alternative with a potentially promising future.

Finally, this study analyses the performance of the different versions of the algorithm in a game per game basis, and it is clear that in some games the agent performance shows a trend after increasing the population size or the individual length. For instance, in most games the agent benefits from using larger populations, but, in some of them, it works better with fewer individuals. Similarly, a long sequence of actions typically helps finding better solutions, but some games form the exception and RHEA performs better with shorter individual lengths. In general, however, it has been observed that an increase in the population size has a higher impact on the performance than considering a further look ahead (longer individuals).

Therefore, although the general finding is that bigger populations and longer individuals improve the performance of RHEA on average, it should be possible to devise methods that could identify the type of game being played, and employ different (or, maybe, modify dynamically) parameter settings. In a form of a meta-heuristic, an agent could be able to select which configuration better fits the game being played at the moment and increases the average performance in this domain.

The most straightforward line of future work, however, is the improvement of the vanilla RHEA in this general setting. The objectives are twofold: first, seeking bigger improvements of action sequences during the evolution phase, without the need of having too broad an exploration as in the case of RS; and second, being able to better handle long individual lengths in order for them to not hinder the evolutionary process. Additionally, further analysis could be conducted on stochastic games, considering the effects of more elite members in the population or resampling individuals, in order to alleviate the effect of noise in the evaluations.

\bibliographystyle{splncs03}
\bibliography{gaina}

\end{document}